  \documentclass[11pt]{article}
  
  \usepackage{cite}
  \usepackage{caption}
  \usepackage{subcaption} 
  \usepackage{graphicx}
  \usepackage{amsmath}
  \usepackage{algorithmic}
  \usepackage{array}
  \usepackage{url}
  \usepackage{tabularx} 
  \usepackage{tabu} 
  \usepackage{multirow}
  \usepackage{longtable} 
  \usepackage{authblk}
  
  \providecommand{\keywords}[1]{\textbf{\textit{Keywords: }} #1}

\begin{document}
  
  \title{A Comparison of Handcrafted and Deep Neural Network Feature Extraction for Classifying Optical Coherence Tomography (OCT) Images}
  
  
  \author{Kuntoro Adi Nugroho}
  \affil{Department of Computer Engineering\\
  Universitas Diponegoro\\
  Semarang, Indonesia\\
  Email: kuntoro@live.undip.ac.id}
  
  \maketitle
  
  \begin{abstract}
  Optical Coherence Tomography allows ophthalmologist to obtain cross-section imaging of eye retina. Assisted with digital image analysis methods, effective disease detection could be performed. Various  methods exist to extract feature from OCT images. The proposed study aims to compare the effectiveness of handcrafted and deep neural network features. The evaluated dataset consist of 32339 instances distributed in four classes, namely CNV, DME, DRUSEN, and NORMAL. The methods are Histogram of Oriented Gradient (HOG), Local Binary Pattern (LBP), DenseNet-169, and ResNet50. As a result, the deep neural network based methods outperformed the handcrafted feature with 88\% and 89\% accuracy for DenseNet and ResNet compared to 50 \% and 42 \% for HOG and LBP respectively. The deep neural network based methods also demonstrated better result on the under represented class.
  \end{abstract}

  
  \section{Introduction}
  
  As a breakthrough of medical imaging technology, with most application in ophthalmology, optical coherence tomography (OCT) allows in situ morphology assessment without invasive procedure (surgical or removal of tissue sample). The OCT imaging technique is based on light, in contrast to ultrasound B-mode imaging that utilizes sound, thus allows more accurate resolution of $10 \mu m$  \cite{fujimoto_optical_1995}. This method is superior to fluorescein angiography in imaging all layers of the retinal vasculature as reported by Spaide et al \cite{spaide_retinal_2015}.  
  
  Digital image processing and analysis tasks in optical coherence tomography include automatic image segmentation and detection of disease. Both tasks are related given that segmentation can be viewed as a recognition task in an $ m x n $ pixels image. Various method are available to extract feature from OCT images. The features can be  handcrafted or obtained from automatic feature learning such as deep neural network. 
  
  Some example of studies on OCT image segmentation based handcrafted features are the study by Vermeer et al. and Lang et al \cite{vermeer_automated_2011} \cite{lang_retinal_2013}. Vermeer et al. proposed  two types of features. The first is pixel value and the pixel above and below it. The second is haar-like feature based on A-line scan \cite{vermeer_automated_2011}. Two general categories of features are introduced by Lang et al. namely spatial aware and context aware features which constitutes 27 features in total \cite{lang_retinal_2013}. Other study by Gadde et al. demonstrated foveal avascular zone (FAZ) produce segmentation using local fractal analysis \cite{gadde_quantification_2016}. 
  
  \begin{figure}[t]
  
    \centering
    \begin{subfigure}[t]{0.22\textwidth}
    \centering
    \includegraphics[width=.95\linewidth]{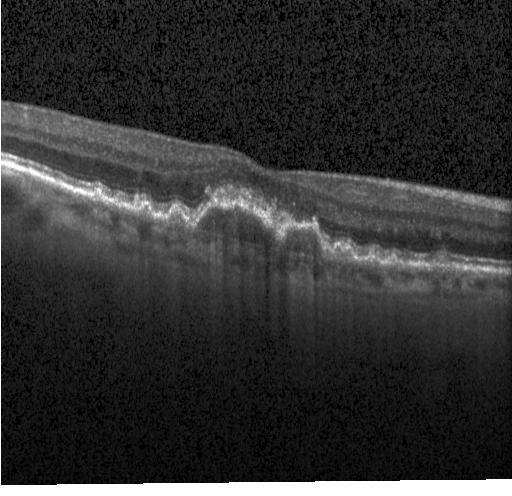}
    \caption{CNV}
    \end{subfigure}%
    ~
    \begin{subfigure}[t]{0.22\textwidth}
    \centering
    \includegraphics[width=.95\linewidth]{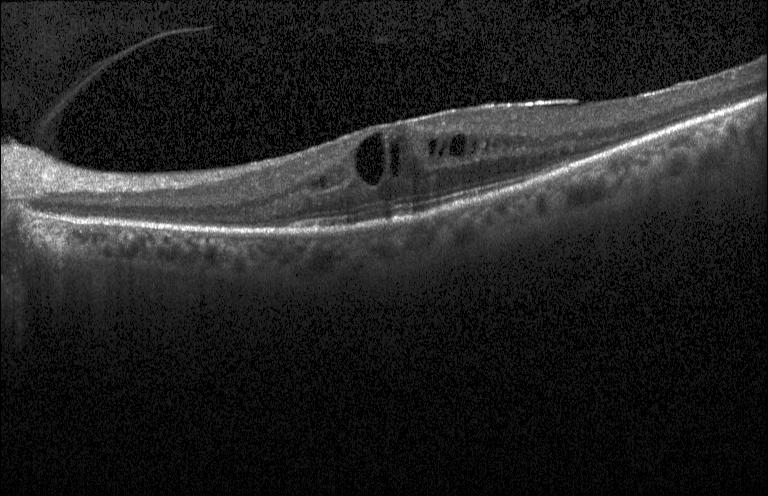}
    \caption{DME}
    \end{subfigure}%
    ~   
    \\
    \begin{subfigure}[t]{0.22\textwidth}
    \centering
    \includegraphics[width=.95\linewidth]{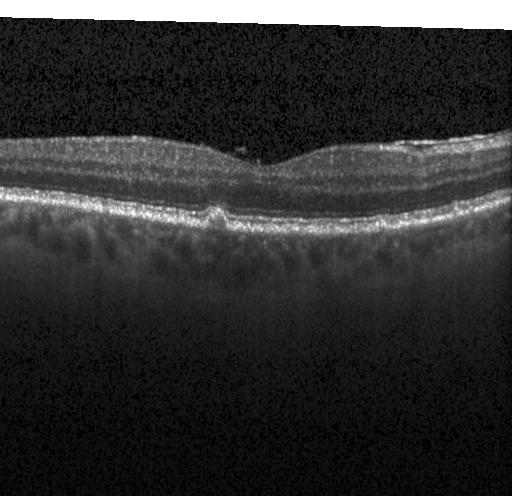}
    \caption{DRUSEN}
    \end{subfigure}%
    ~
    \begin{subfigure}[t]{0.22\textwidth}
    \centering
    \includegraphics[width=.95\linewidth]{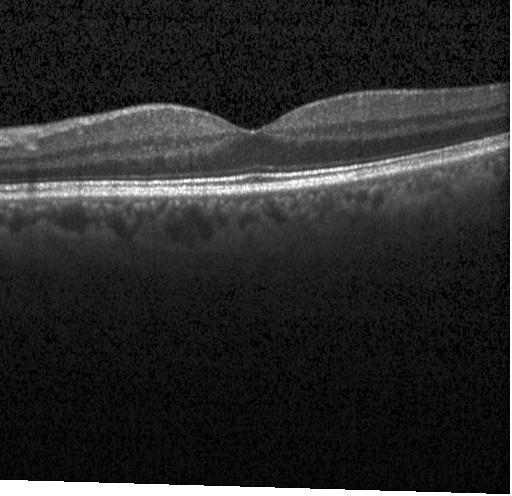}
    \caption{NORMAL}
    \end{subfigure}%

    \caption{Dataset samples \cite{kermany_identifying_2018}}
    \label{fig:datasetSample}    
    
  \end{figure}
  
  Handcrafted features work relatively well in specific domain with small number of data. With continuous improvements, feature learning based on deep neural network is more favorable due to possibility of learning from larger data and to learn more complex pattern. This is achieved by constructing the network with deeper layer. Learning from small number of sample is also possible with transfer learning.
  
  OCT segmentation with automatic feature extraction are demonstrated in the study by He et al. and Alonso-Caneiro et al \cite{he_topology_2018} \cite{alonso-caneiro_automatic_nodate}. U-Net based Segmentation Network (S-Net) followed by Regression Network (R-Net) were proposed to obtain topology guaranteed segmentation \cite{he_topology_2018}. In contrast to the S-Net and R-Net framework, Convolutional Neural Network (CNN) was trained to compute probability map as a sign of the presence of boundary at certain pixels \cite{alonso-caneiro_automatic_nodate}. Both studies demonstrated the superiority of automatic feature extraction compared to the non-automatic counterpart.
  
  Coherent with the presented OCT segmentation studies, automatically and non-automatically extracted features are also evaluated in OCT image classification studies. 
  
  Atlas based shape analysis with Linear Discriminant Analysis (LDA) was studied by Lee et al. The method was evaluated for two experiments, the first was discerning glaucoma versus normal eye and the second was normal versus suspect eye (normal eye in pair with unilateral glaucoma). The proposed method relies on fshape framework to obtain atlas estimation, a template object from which each observation based on. With reliance on initial template to compute the final mean template, the method requires a good initial template to start with. The study was also limited in terms of the number of evaluated data \cite{lee_atlas-based_2017}. Other reported study in OCT is Histogram of Oriented Gradient (HOG) feature extractor and Support Vector Machine for classifying normal, diabetic macular edema (DME), and dry age-related macular degeneration (AMD). Evaluated on spectral domain-OCT data from 45 patients, the correctly classified fraction of volumes are $ 86.67\%, 100\%, $ and $ 100\% $ respectively \cite{srinivasan_fully_2014}. 
  
  Traditional computer vision application in object recognition usually employs manually extracted feature. With deep neural network feature extraction, limitation on the use of good initial template and relatively small number of data sample as required in the previously discussed studies could be mitigated. 
  
  Convolutional neural network has been studied at least before 1995 with application in hand written word recognition \cite{bengio_globally_nodate}. A notable success of this method is in ImageNet object recognition \cite{krizhevsky_imagenet_2017}. 
  
  The success has been followed up in medical image analysis including application in fundus autofluorescence (FAF) and OCT images. In FAF, for example, deep convolutional neural network (DCNN) performed $96\%$ and $91\%$ in recognizing geographic atrophy (GA) versus normal and GA vs other diseases respectively. Reported study in OCT image includes the proposed VGG-16 with Xavier algorithm weight initialization which successfully identify age-related macular degeneration with ROC area under curve of $ 92.78 \% $ and accuracy of $ 87.63 \% $ \cite{lee_deep_2017}. Other study is the identification and quantification of intraretinal cystoid fluid (IRC) and subretinal fluid (SRF) that resulted in a high correlation between the method and the ground truth \cite{schlegl_fully_2018}.  
  
  In spite of the recent development in deep neural network based methods, comparison and evaluation of deep neural network and non-automatic (handcrafted) feature extraction in OCT has not been thoroughly reported. Therefore, the proposed study aims to evaluate image feature extractions methods for classifying OCT images. Two types of methods are considered. The first method belongs to the non-automatic feature extraction, while the second belongs to deep convolutional neural network family.
  
  \section{Research Method}
  
  \subsection{Dataset}
  
  OCT images from the study by Kermany et al. are used in this study \cite{kermany_identifying_2018}. The OCT images belong to four class, namely Choroidal Neovascularization (CNV), Diabetic Macular Edema (DME), Drusen (presents in eary Age-Related Macular Degeneration), and Normal. Some of the samples are shown in Fig \ref{fig:datasetSample}. The data distribution is shown in Table \ref{tab:datasetDistribution}.
  
  \begin{table}[h!]
    \begin{center}
      \caption{Dataset Instance Distribution}
      \label{tab:datasetDistribution}
      \begin{tabular}{|l|c c c|r|} 
       \hline 
        \textbf{ } & \textbf{train} & \textbf{test} & \textbf{val} & \\
        \hline
        \textbf{CNV} & 37205 & 242 & 8 & 37455 \\
        \textbf{DME} & 11348 & 242 & 8 & 11598 \\
        \textbf{DRUSEN} & 8616 & 242 & 8 & 8866 \\
        \textbf{NORMAL} & 26315 & 242 & 8 & 26565 \\
        \hline
        \textit{ } & 83484 & 968 & 32 & 84484 \\
        \hline	
      \end{tabular}
    \end{center}
  \end{table}

  \subsection{Methodology}
  
  The proposed methodology consists of the following steps
  
  \begin{enumerate}
  
  \item Data preprocessing
  
  \item Feature extraction
  
  \item Classifier training
  
  \end{enumerate}
  
  \subsection{Data Preprocessing}
  
  \begin{enumerate}
  
  \item Image rescaling and padding
  
  The original images are varied in size. In order to extract the feature, all of the images were adjusted into uniform size of 224 x 224 pixels. The longer dimension (between height or width) will be rescaled into 224, and the shorter will be adjusted with maintaining the image aspect ratio. After rescaling, the image was padded with zero pixels located in the center. The procedure is illustrated in Fig \ref{fig:rescalePad}.
  
  \begin{figure}
  \centering
  \begin{subfigure}[t]{0.35\textwidth}
    \centering
    \includegraphics[width=.8\linewidth]{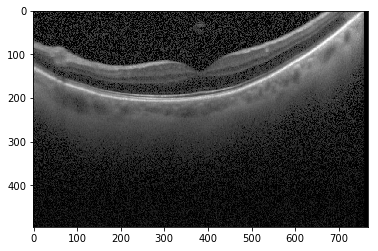}
    \caption{}
    \end{subfigure}%
    ~
  \begin{subfigure}[t]{0.15\textwidth}
    \centering
    \includegraphics[width=.8\linewidth]{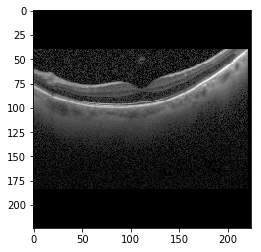}
    \caption{}
    \end{subfigure}%
  \caption{Rescaling and padding process: A single channel 469 x 768  image (a) was rescaled into 144 x 224 and then padded with black pixels to form 224 x 224 pixels image (b)}
  \label{fig:rescalePad}
  \end{figure}
   
  \item Image Resampling
  
  There is a large gap between the number of training samples (83484) and  validation samples (32). Given the relatively large number of instances and the imbalance between training and validation data as shown in Table \ref{tab:datasetDistribution}, the training and validation data were resampled and re-distributed. Approximately $ \pm 25 \% $ of the training images were randomly sampled to form the new training set. Another $ \pm 12.5 \% $ different partition of random samples were taken from training set for the new validation set. The test set was not altered. Therefore, only 38.278 \% of the data were retained. The new distribution is show in Table \ref{tab:newDistribution}. 
  
  \end{enumerate}
  
  \begin{table}[h!]
    \begin{center}
      \caption{Dataset Instance Distribution After Resampling}
      \label{tab:newDistribution}
      \begin{tabular}{|l|c c c|r|} 
       \hline 
        \textbf{ } & \textbf{train} & \textbf{test} & \textbf{val} & \\
        \hline
        \textbf{CNV} & 9261 & 242 & 4743 & 14246 \\
        \textbf{DME} & 2803 & 242 & 1441 & 4486 \\
        \textbf{DRUSEN} & 2106 & 242 & 1096 & 3444 \\
        \textbf{NORMAL} & 6571 & 242 & 3350 & 10163 \\
        \hline
        \textit{ } & 20741 & 968 & 10630 &  32339\\
        \hline	
      \end{tabular}
    \end{center}
  \end{table}

  \subsection{Feature Extraction}
  
  The training, validation, and testing images were converted into feature vector using the following methods
  
  \begin{enumerate}
  
  \item Histogram of Oriented Gradient (HOG) \cite{dalal_histograms_2005}
  
  \item Local Binary Pattern (LBP) \cite{ojala_performance_1994} \cite{ojala_multiresolution_2002}
  
  \item Pretrained Convolutional Neural Network: Residual Network (ResNet50) \cite{he_deep_2016}
  
  \item Pretrained Convolutional Neural Network: Densely Connected Network (DenseNet-169) \cite{huang_densely_2016}
  
  \end{enumerate}
  
  \begin{figure}[t!]
    \centering
    
    \begin{subfigure}[t]{0.25\textwidth}
    \centering
    \includegraphics[width=.8\linewidth]{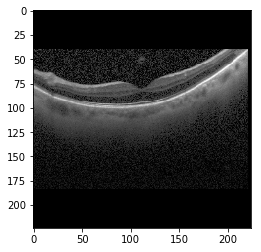}
    \caption{Image sample}
    \end{subfigure}%
    ~
    \\
    \begin{subfigure}[t]{0.25\textwidth}
    \centering
    \includegraphics[width=.8\linewidth]{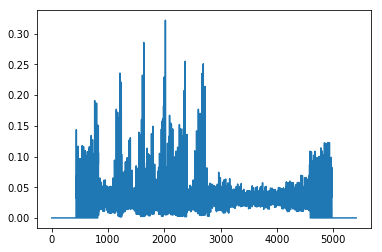}
    \caption{HOG feature}
    \end{subfigure}%
    ~
    \begin{subfigure}[t]{0.25\textwidth}
    \centering
    \includegraphics[width=.8\linewidth]{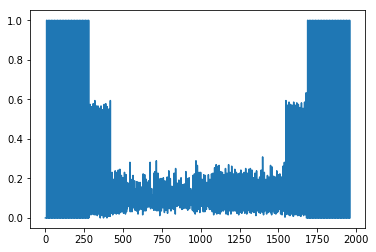}
    \caption{LBP feature}
    \end{subfigure}%
    ~   
    \\
    \begin{subfigure}[t]{0.25\textwidth}
    \centering
    \includegraphics[width=.8\linewidth]{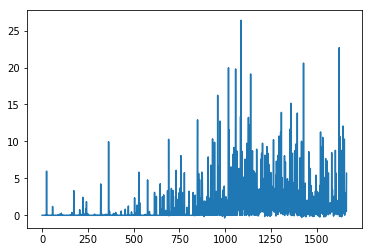}
    \caption{DenseNet-169 feature}
    \end{subfigure}%
    ~
    \begin{subfigure}[t]{0.25\textwidth}
    \centering
    \includegraphics[width=.8\linewidth]{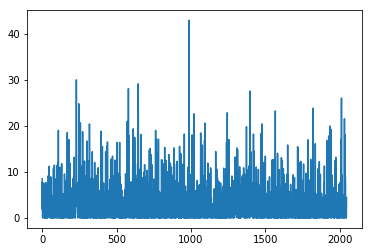}
    \caption{ResNet50 feature}
    \end{subfigure}%
    \caption{Feature extraction example (image: normal-336530-5.png)}
    
  \end{figure}
  
  HOG and LBP belong to non-automatic feature extraction method, while the other two convolutional neural networks (ResNet50 and DenseNet-169) belong to the automatic method. The parameter settings for HOG and LBP are summarized in Table \ref{tab:hoglbpSettings}. 
  
  Both of the convolutional neural network feature extractors are based on Keras implementation \cite{chollet_keras_2015}. The network parameter and size is described in Table \ref{tab:networkSize}. The weights are obtained by pretraining on Imagenet data.

  \begin{table}[h!]
    \begin{center}
      \caption{HOG and LBP Parameter Settings}
      \label{tab:hoglbpSettings}
      \begin{tabular}{| l |l|c|} 
       \hline 
        \textbf{ } & \textbf{Parameter} & \textbf{Value} \\
        \hline
        \multirow{3}{*}{\textbf{HOG}} & Number of orientations & 8 \\
         & Pixels per cell & 16 x 16 \\
         & Cells per block & 2 x 2 \\ \hline
        \multirow{3}{*}{\textbf{LBP}} & Number of Points & 16 \\
        & Radius & 2 \\
        & Histogram block size & 16 x 16 \\
         & Method & Uniform \\
        \hline	
      \end{tabular}
    \end{center}
  \end{table}
  
  \begin{table}[h!]
    \begin{center}
      \caption{Network Size}
      \label{tab:networkSize}
      \begin{tabular}{| l | c | c |} 
       \hline 
        \textbf{ } & \textbf{Number of Parameters} & \textbf{Size on Disk} \\
        \hline
         DenseNet-169 & 14,307,880 & 57 MB \\
         ResNet50 & 25,636,712 & 99 MB \\
        \hline	
      \end{tabular}
    \end{center}
  \end{table}
  
  The resulting feature dimension for each method is summarized in Table \ref{tab:featureSize}.
  
  \begin{table}[h!]
    \begin{center}
      \caption{Feature Size}
      \label{tab:featureSize}
      \begin{tabular}{| c | c | c | c |} 
       \hline 
        \textbf{ HOG } & \textbf{LBP} & \textbf{DenseNet} & \textbf{ResNet} \\
        \hline
         5408 & 1960 & 1664 & 2048 \\
        \hline	
      \end{tabular}
    \end{center}
  \end{table}
  
  \subsection{Classification}
  
  The extracted features were then utilized to train a classifier. A classifier was trained for each feature extraction type. The evaluated classifier is a perceptron neural network with no hidden layer (logistic regression). Although multilayer perceptron is a possible option to learn more complex (non-linear) pattern, linear classifier is sufficient for benchmarking. 
  
  The classifier was implemented using one-vs-all strategy. Therefore, four perceptron units were trained for each classifier as there are four classes (CNV, DME, DRUSEN, NORMAL). The classifier uses softmax activation function. 
  
  The training parameters were summarized in Table \ref{tab:trainParams}.
  
  \begin{table}[h!]
    \begin{center}
      \caption{Training Parameters}
      \label{tab:trainParams}
      \begin{tabular}{| l | c |} 
       \hline 
        \textbf{Parameter} & \textbf{Value} \\
        \hline
         Number of epochs & 100 \\
         Learning rate & $ 10^{-4} $ \\
         Optimizer & Adam \cite{kingma_adam:_2014} \\
         Loss function & Categorical crossentropy \\
        \hline	
      \end{tabular}
    \end{center}
  \end{table}
  
  \section{Result and Discussion}
  
  \subsection{Result}
  
  \subsubsection{Training and Validation}
  
\begin{figure}[t]
  
    \centering
    \begin{subfigure}[t]{0.25\textwidth}
    \centering
    \includegraphics[width=.95\linewidth]{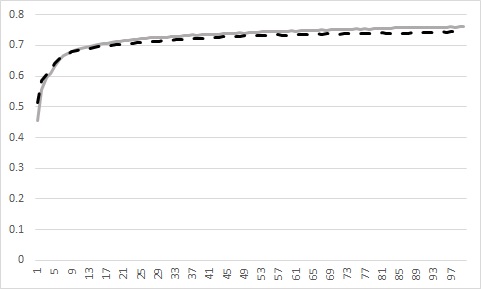}
    \caption{HOG}
    \end{subfigure}%
    ~
    \begin{subfigure}[t]{0.25\textwidth}
    \centering
    \includegraphics[width=.95\linewidth]{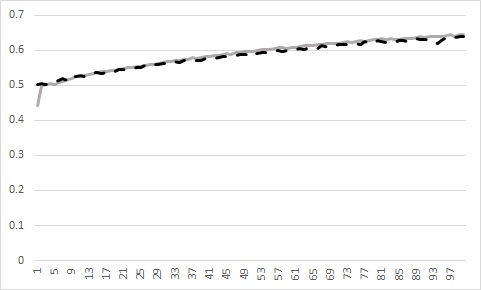}
    \caption{LBP}
    \end{subfigure}%
    ~   
    \\
    \begin{subfigure}[t]{0.25\textwidth}
    \centering
    \includegraphics[width=.95\linewidth]{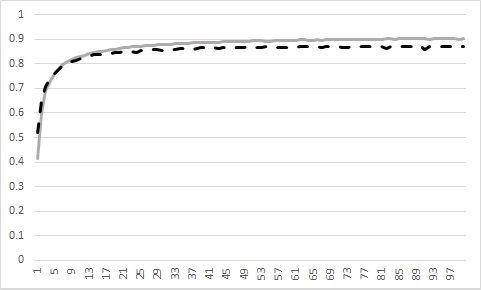}
    \caption{DenseNet-169}
    \end{subfigure}%
    ~
    \begin{subfigure}[t]{0.25\textwidth}
    \centering
    \includegraphics[width=.95\linewidth]{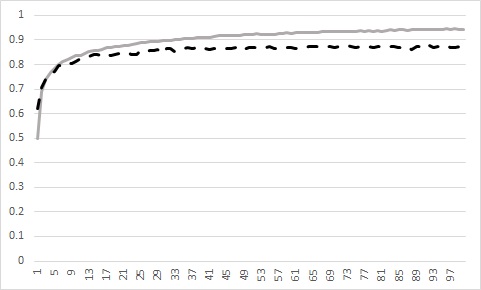}
    \caption{ResNet50}
    \end{subfigure}%

    \caption{Training and Validation Accuracy in 100 Epochs. The gray straight line and the black dashed line denote training and validation accuracy respectively.}
    \label{fig:trainValAcc}    
    
  \end{figure}

\begin{figure}
    \centering
      \includegraphics[width=0.45\textwidth]{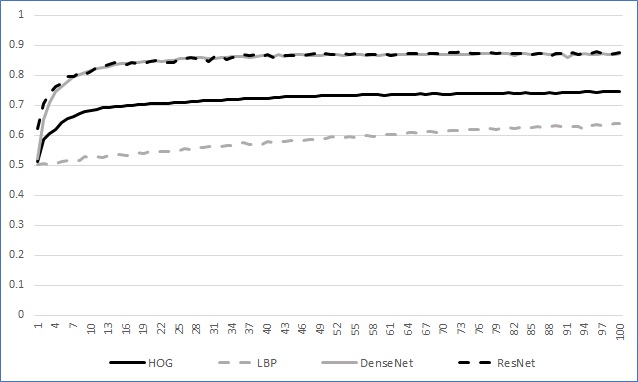}
    \caption{Validation accuracy for each method against each other}
    \label{fig:valAcc}
  \end{figure}

  Fig \ref{fig:trainValAcc} shows per-epoch training and validation accuracy for each method. The validation accuracy of the four methods against each other is depicted in Fig \ref{fig:valAcc}.
  
  \subsubsection{Evaluation in Test Data}
  
  Evaluated on test data, the classifier accuracy is shown in Table \ref{tab:testAccuracy}. The precision, recall, and F1-score for HOG, LBP, DenseNet, and ResNet are summarized in Table \ref{tab:evaluationHOG}, \ref{tab:evaluationLBP}, \ref{tab:evaluationDense}, and \ref{tab:evaluationResidual} respectively.
  
  \begin{table}[h!]
    \begin{center}
      \caption{Classifier with HOG Feature}
      \label{tab:evaluationHOG}
      \begin{tabular}{|l|c c c|} 
       \hline 
        \textbf{ } & \textbf{Precision} & \textbf{Recall} & \textbf{F1-score} \\
        \hline
        \textbf{CNV} & 0.40 & 0.96 & 0.57 \\
        \textbf{DME} & 0.60 & 0.04 & 0.07 \\
        \textbf{DRUSEN} & 0.59 & 0.04 & 0.08  \\
        \textbf{NORMAL} & 0.66 & 0.96 & 0.78 \\
        \hline
        \textit{ average } & 0.56 & 0.50 & 0.37 \\
        \hline	
      \end{tabular}
    \end{center}
  \end{table}
  
  \begin{table}[h!]
    \begin{center}
      \caption{Classifier with LBP Feature}
      \label{tab:evaluationLBP}
      \begin{tabular}{|l|c c c|} 
       \hline 
        \textbf{ } & \textbf{Precision} & \textbf{Recall} & \textbf{F1-score} \\
        \hline
        \textbf{CNV} & 0.36 & 0.95 & 0.52 \\
        \textbf{DME} & 0.00 & 0.00 & 0.00 \\
        \textbf{DRUSEN} & 0.00 & 0.00 & 0.00 \\
        \textbf{NORMAL} & 0.55 & 0.74 & 0.63 \\
        \hline
        \textit{ average } & 0.23 & 0.42 & 0.29 \\
        \hline	
      \end{tabular}
    \end{center}
  \end{table}
  
  \begin{table}[h!]
    \begin{center}
      \caption{Classifier with DenseNet-169 Feature}
      \label{tab:evaluationDense}
      \begin{tabular}{|l|c c c|} 
       \hline 
        \textbf{ } & \textbf{Precision} & \textbf{Recall} & \textbf{F1-score} \\
        \hline
        \textbf{CNV} & 0.72 & 0.99 & 0.84 \\
        \textbf{DME} & 0.98 & 0.79 & 0.87 \\
        \textbf{DRUSEN} & 0.98 & 0.76 & 0.86 \\
        \textbf{NORMAL} & 0.94 & 0.98 & 0.96 \\
        \hline
        \textit{ average } & 0.90 & 0.88 & 0.88 \\
        \hline	
      \end{tabular}
    \end{center}
  \end{table}
  
  \begin{table}[h!]
    \begin{center}
      \caption{Classifier with ResNet50 Feature}
      \label{tab:evaluationResidual}
      \begin{tabular}{|l|c c c|} 
       \hline 
        \textbf{ } & \textbf{Precision} & \textbf{Recall} & \textbf{F1-score} \\
        \hline
        \textbf{CNV} & 0.75 & 1.00 & 0.86 \\
        \textbf{DME} & 1.0 & 0.80 & 0.89 \\
        \textbf{DRUSEN} & 0.95 & 0.79 & 0.86 \\
        \textbf{NORMAL} & 0.94 & 0.98 & 0.96 \\
        \hline
        \textit{ average } & 0.91 & 0.89 & 0.89 \\
        \hline	
      \end{tabular}
    \end{center}
  \end{table}
  
  \begin{figure}
    \centering
      \includegraphics[width=0.45\textwidth]{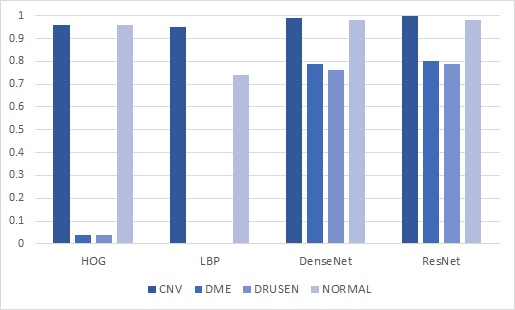}
    \caption{Recall Across The Four Methods}
    \label{fig:recallPlot}
  \end{figure}
  
  \begin{table}[h!]
    \begin{center}
      \caption{Test Accuracy}
      \label{tab:testAccuracy}
      \begin{tabular}{| c | c | c | c |} 
       \hline 
        \textbf{ HOG } & \textbf{LBP} & \textbf{DenseNet} & \textbf{ResNet} \\
        \hline
         0.5010330 & 0.4235537 & 0.880165 & 0.8925619 \\
        \hline	
      \end{tabular}
    \end{center}
  \end{table}
  
  \subsection{Discussion}

  Fig \ref{fig:trainValAcc} shows that the training and validation accuracy converges by the end of the training process. The progress is almost static at the final epochs, although LBP still shows relatively higher gradient compared to other methods. 
  
  HOG and LBP show almost no gap between training and validation accuracy, while a larger gap is shown by the CNN based methods, especially ResNet50. The gap is related to the bias-variance characteristic of the methods. The larger the gap, the more the model variance. Against each other, the CNN based methods consistently demonstrated higher accuracy than HOG and LBP, as shown in Fig \ref{fig:valAcc}. As shown by the lower error (bias on training and validation data), the CNN based method produced better feature compared to HOG and LBP.    
  
  The performance on the test data is consistent with the performance on the training data. Overall, ResNet and DenseNet features outperformed HOG and LBP in terms of accuracy, as shown in Table \ref{tab:testAccuracy}. The rank for each class recall is almost consistent across the methods, as visualized in Fig \ref{fig:recallPlot}. The recall on CNV is the highest for all methods (0.96, 0.95, 0.99, 1.0 for each HOG, LBP, DenseNet, and ResNet), followed by the recall on NORMAL. The result is very likely due to the distribution of the data. CNV and NORMAL class are the classes with the larger number of instances compared to the rest, as shown in Table \ref{tab:datasetDistribution}. Although DME and DRUSEN slightly under represented, DenseNet and ResNet still performed much higher than other methods, greater than 75 \% in terms of recall.
  
  In terms of classification performance, ResNet50 slightly outperformed DenseNet-169. However, considering the feature and network parameters, DenseNet-169 performed almost equally with ResNet50 with much less parameter and network size ($\pm$ 14 M compared to $\pm$ 25 M, see Table \ref{tab:networkSize}).

  \section{Conclusion}
  
  In conclusion, the classifiers trained on features extracted from deep neural network demonstrated the best performance. The result indicates that deep neural network based methods generates better feature compared to HOG and LBP in OCT image classification.   
  
  \bibliographystyle{unsrt}
  \bibliography{references}

\begin{thebibliography}{10}

\bibitem{fujimoto_optical_1995}
James~G Fujimoto, Mark~E Brezinski, Guillermo~J Tearney, Stephen~A Boppart,
  Brett Bouma, Michael~R Hee, James~F Southern, and Eric~A Swanson.
\newblock Optical biopsy and imaging using optical coherence tomography.
\newblock 1(9):970--972.

\bibitem{spaide_retinal_2015}
Richard~F. Spaide, James~M. Klancnik, and Michael~J. Cooney.
\newblock Retinal vascular layers imaged by fluorescein angiography and optical
  coherence tomography angiography.
\newblock 133(1):45.

\bibitem{vermeer_automated_2011}
K.~A. Vermeer, J.~van~der Schoot, H.~G. Lemij, and J.~F. de~Boer.
\newblock Automated segmentation by pixel classification of retinal layers in
  ophthalmic {OCT} images.
\newblock 2(6):1743.

\bibitem{lang_retinal_2013}
Andrew Lang, Aaron Carass, Matthew Hauser, Elias~S. Sotirchos, Peter~A.
  Calabresi, Howard~S. Ying, and Jerry~L. Prince.
\newblock Retinal layer segmentation of macular {OCT} images using boundary
  classification.
\newblock 4(7):1133.

\bibitem{gadde_quantification_2016}
Santosh G.~K. Gadde, Neha Anegondi, Devanshi Bhanushali, Lavanya Chidambara,
  Naresh~Kumar Yadav, Aruj Khurana, and Abhijit Sinha~Roy.
\newblock Quantification of vessel density in retinal optical coherence
  tomography angiography images using local fractal dimension.
\newblock 57(1):246.

\bibitem{kermany_identifying_2018}
Daniel~S Kermany, Michael Goldbaum, Wenjia Cai, Carolina~CS Valentim, Huiying
  Liang, Sally~L Baxter, Alex McKeown, Ge~Yang, Xiaokang Wu, Fangbing Yan, and
  {others}.
\newblock Identifying medical diagnoses and treatable diseases by image-based
  deep learning.
\newblock 172(5):1122--1131.

\bibitem{he_topology_2018}
Yufan He, Aaron Carass, Bruno~M. Jedynak, Sharon~D. Solomon, Shiv Saidha,
  Peter~A. Calabresi, and Jerry~L. Prince.
\newblock Topology guaranteed segmentation of the human retina from {OCT} using
  convolutional neural networks.

\bibitem{alonso-caneiro_automatic_nodate}
David Alonso-Caneiro, Scott~A Read, Jared Hamwood, Stephen~J Vincent, and
  Michael~J Collins.
\newblock Automatic segmentation of retinal and choroidal thickness in {OCT}
  images using convolutional neural networks.
\newblock page~1.

\bibitem{lee_atlas-based_2017}
Sieun Lee, Nicolas Charon, Benjamin Charlier, Karteek Popuri, Evgeniy Lebed,
  Marinko~V. Sarunic, Alain Trouvé, and Mirza~Faisal Beg.
\newblock Atlas-based shape analysis and classification of retinal optical
  coherence tomography images using the functional shape (fshape) framework.
\newblock 35:570--581.

\bibitem{srinivasan_fully_2014}
Pratul~P. Srinivasan, Leo~A. Kim, Priyatham~S. Mettu, Scott~W. Cousins,
  Grant~M. Comer, Joseph~A. Izatt, and Sina Farsiu.
\newblock Fully automated detection of diabetic macular edema and dry
  age-related macular degeneration from optical coherence tomography images.
\newblock 5(10):3568.

\bibitem{bengio_globally_nodate}
Yoshua Bengio, Yann {LeCun}, and Donnie Henderson.
\newblock Globally trained handwritten word recognizer using spatial
  representation, convolutional neural networks, and hidden markov models.
\newblock page~8.

\bibitem{krizhevsky_imagenet_2017}
Alex Krizhevsky, Ilya Sutskever, and Geoffrey~E. Hinton.
\newblock {ImageNet} classification with deep convolutional neural networks.
\newblock 60(6):84--90.

\bibitem{lee_deep_2017}
Cecilia~S. Lee, Doug~M. Baughman, and Aaron~Y. Lee.
\newblock Deep learning is effective for classifying normal versus age-related
  macular degeneration {OCT} images.
\newblock 1(4):322--327.

\bibitem{schlegl_fully_2018}
Thomas Schlegl, Sebastian~M. Waldstein, Hrvoje Bogunovic, Franz Endstraßer,
  Amir Sadeghipour, Ana-Maria Philip, Dominika Podkowinski, Bianca~S. Gerendas,
  Georg Langs, and Ursula Schmidt-Erfurth.
\newblock Fully automated detection and quantification of macular fluid in
  {OCT} using deep learning.
\newblock 125(4):549--558.

\bibitem{dalal_histograms_2005}
N.~Dalal and B.~Triggs.
\newblock Histograms of oriented gradients for human detection.
\newblock In {\em 2005 {IEEE} Computer Society Conference on Computer Vision
  and Pattern Recognition ({CVPR}'05)}, volume~1, pages 886--893. {IEEE}.

\bibitem{ojala_performance_1994}
T.~Ojala, M.~Pietikainen, and D.~Harwood.
\newblock Performance evaluation of texture measures with classification based
  on kullback discrimination of distributions.
\newblock In {\em Proceedings of 12th International Conference on Pattern
  Recognition}, volume~1, pages 582--585. {IEEE} Comput. Soc. Press.

\bibitem{ojala_multiresolution_2002}
T.~Ojala, M.~Pietikainen, and T.~Maenpaa.
\newblock Multiresolution gray-scale and rotation invariant texture
  classification with local binary patterns.
\newblock 24(7):971--987.

\bibitem{he_deep_2016}
Kaiming He, Xiangyu Zhang, Shaoqing Ren, and Jian Sun.
\newblock Deep residual learning for image recognition.
\newblock In {\em 2016 {IEEE} Conference on Computer Vision and Pattern
  Recognition ({CVPR})}, pages 770--778. {IEEE}.

\bibitem{huang_densely_2016}
Gao Huang, Zhuang Liu, Laurens van~der Maaten, and Kilian~Q. Weinberger.
\newblock Densely connected convolutional networks.

\bibitem{chollet_keras_2015}
François Chollet and {others}.
\newblock {\em Keras}.

\bibitem{kingma_adam:_2014}
Diederik~P. Kingma and Jimmy Ba.
\newblock Adam: A method for stochastic optimization.

\end{thebibliography}

  \end{document}